\documentclass[letterpaper, 10 pt, conference]{ieeeconf}  

\IEEEoverridecommandlockouts
\let\labelindent\relax

\usepackage{amsmath,amssymb,amsthm,mathtools,bm}
\usepackage{graphicx}
\usepackage{xcolor}
\usepackage{enumitem}
\usepackage{booktabs}
\usepackage{algorithm}
\usepackage{algorithmic}
\usepackage{cite}
\usepackage[hidelinks]{hyperref}
\usepackage{caption}

\newcommand{\incfig}[2][]{\includegraphics[#1]{\detokenize{#2}}}

\providecommand{\citep}[1]{\cite{#1}}
\providecommand{\citet}[1]{\cite{#1}}

\DeclareMathOperator{\tr}{tr}
\newcommand{\R}{\mathbb{R}}

\theoremstyle{plain}

\theoremstyle{definition}

\title{Benchmarking Reinforcement Learning via Stochastic Converse Optimality: Generating Systems with Known Optimal Policies}

\author{%
Sinan Ibrahim$^{1}$, Gr\'egoire Ouerdane$^{1}$, Hadi Salloum$^{2}$, Henni Ouerdane$^{1}$, Stefan Streif$^{3}$, Pavel Osinenko$^{1,4,5}$%
\thanks{$^{1}$Skolkovo Institute of Science and Technology, Moscow, Russia. Email: \{sinan.ibrahim, gregoire.ouerdane, h.ouerdane, p.osinenko\}@skoltech.ru}%
\thanks{$^{2}$Innopolis University, Innopolis, Russia. Email: h.salloum@innopolis.ru}%
\thanks{$^{3}$Technische Universit\"at Chemnitz, Chemnitz, Germany. Email: stefan.streif@etit.tu-chemnitz.de}%
\thanks{$^{4}$Central University, Moscow, Russia}%
\thanks{$^{5}$Sirius University of Science and Technology, Sotchi, Russia}%
\thanks{Corresponding author: Pavel Osinenko, p.osinenko@gmail.com.}%
}

\begin{document}
\maketitle
\thispagestyle{empty}
\pagestyle{empty}

\begin{abstract}
The objective comparison of Reinforcement Learning (RL) algorithms
is notoriously complex as outcomes and benchmarking of performances of different RL approaches are critically sensitive to environmental design, reward structures, and stochasticity inherent in both algorithmic learning and environmental dynamics. To manage this complexity, we introduce a rigorous benchmarking framework by extending converse optimality to discrete-time, control-affine, nonlinear systems with noise. Our framework provides necessary and sufficient conditions, under which a prescribed value function and policy are optimal for constructed systems, enabling the systematic generation of benchmark families via homotopy variations and randomized parameters. We validate it by automatically constructing diverse environments, demonstrating our framework’s capacity for a controlled and comprehensive evaluation across algorithms. By assessing standard methods against a ground-truth optimum, our work delivers a reproducible foundation for precise and rigorous RL benchmarking.
\end{abstract}


\section{Introduction}

The design of optimal feedback controllers for nonlinear systems is governed by the Hamilton-Jacobi-Bellman (HJB) equation, which is computationally intractable for all but the lowest-dimensional systems due to the curse of dimensionality \citep{Bellman1957}. This problem has stimulated the development of alternative, suboptimal methods like reinforcement learning (RL), that trade global optimality for reduced computational complexity. However, without knowledge of true optimal solutions, there is no systematic methodology for testing and evaluating these alternative approaches \citep{nevistic1996constrained}. The ability to benchmark and verify advanced control algorithms used in RL is complicated by a lack of high-dimensional nonlinear test problems with known optimal solutions. Moreover, the difficulty is increased by the strong dependence of results on reward shaping, the choice of system dynamics, stochasticity, and evaluation protocols \citep{Henderson2018DeepRLThatMatters, Agarwal2021Precipice}. Existing benchmarks aggregate scores over heterogeneous suites \citep{Cobbe2020Procgen, Fu2020D4RL}, true optima remaining unavailable for a reliable estimation of optimality gaps, characterization of sample efficiency, and diagnosis of failure modes.

Converse optimality addresses this need by solving the converse problem instead of the HJB equation: given a target optimal value function and stage costs, what are the system dynamics and feedback laws for which Bellman’s optimality equation holds globally? This approach has been successfully applied to generate RL benchmarks for deterministic systems to test control methods \citep{lewis2012reinforcement, vrabie2009neural, vamvoudakis2011multi}: converse theorems yield a closed-form characterization of drifts that realize a prescribed value function \citep{gohrt2019converse}. However, a significant gap exists for stochastic systems, which remain unaddressed due to the complexity that noise adds to solving the HJB equation.

The present work builds a new benchmarking pipeline for RL by extending converse optimality to a widely used class of stochastic, discounted, discrete-time, control-affine, non-linear systems. Our contributions: (i) a discounted stochastic converse optimality framework with a theorem for the case of additive Gaussian noise and a Quadratic--Gaussian (QG) corollary; (ii) a constructive benchmark generation via homotopy over control strength; and (iii) a paired evaluation protocol (Common Random Numbers) with baseline implementations, reporting absolute metrics (optimality gaps, regret) against certified optima.

\begin{figure*}[t]
\centering
\incfig[width=\textwidth]{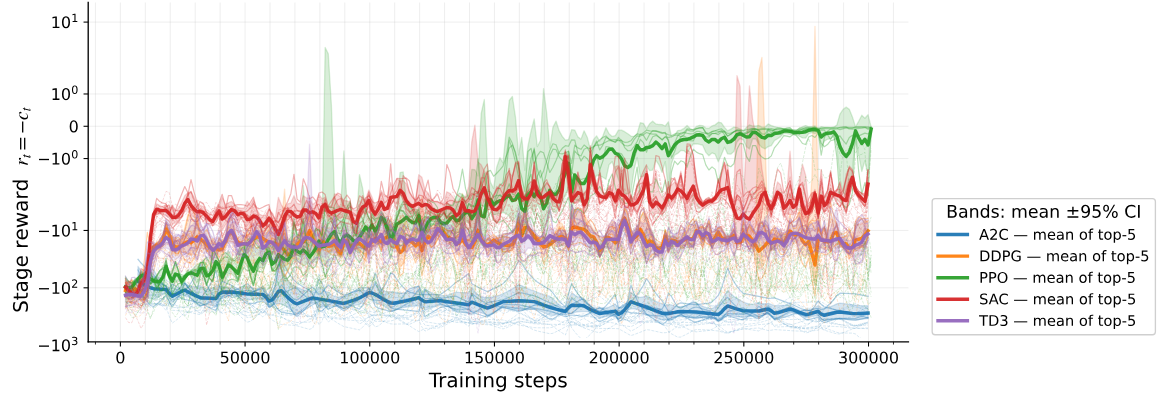}
\caption{Training dynamics snapshot: stage reward $r=-c$ over learning for representative algorithms on a fixed fixture under CRN evaluation. The core comparisons in this paper use absolute, oracle-referenced metrics (OptGap, regret).}
\label{fig:stage_reward}
\end{figure*}

\section{Related Work}

The need for reproducible evaluation in reinforcement learning has
driven the creation of benchmarking tools. Reproducible RL evaluation remains challenging. Henderson et al. \citep{Henderson2018DeepRLThatMatters} showed that the apparent performance of an algorithm is a function of often overlooked experimental conditions, such as hyperparameters and implementation differences \citep{islam2017reproducibility}. This lack of reproducibility
and insufficient statistical rigor hinders progress in RL. In response, newer benchmarks isolate specific facets of this unreliability. The Procgen benchmark confronts the tendency
of models to memorize training environments by using procedural generation \citep{Cobbe2020Procgen}. The D4RL benchmark provides datasets that realistically represent distributional shifts for offline RL \citep{Fu2020D4RL}. A more recent
contribution, Open RL Benchmark \citep{huang2024open}, offers a centralized repository of fully tracked RL experiments. It aggregates over 25,000 runs and documents performance results and experimental conditions. Despite its benefits, Open RL Benchmark faces the same fundamental limitation as its predecessors: the true globally optimal policy is unknown. This makes it impossible to calculate the optimality gap, the most direct performance metric. Without this ground truth, evaluation is necessarily relative and
can be confounded by high variance and sensitivity to protocol
choices \citep{Agarwal2021Precipice, Henderson2018DeepRLThatMatters}.

Converse optimality enables absolute evaluation by constructing problems with certified optimal solutions. Several works have formalized this idea for both discrete-time \citep{gohrt2019converse} and continuous-time \citep{yamerenko2025generating} deterministic settings, establishing
conditions under which a given value function can be viewed as the solution to an optimal control problem. This forms the basis
for a new benchmarking paradigm that replaces cumulative reward
curves with rigorous statistical testing. Our work extends this paradigm to the stochastic setting, using benchmarks representative
of real-world systems to provide a certified optimum and allow an
absolute assessment of the optimality gap.

\section{Problem Setup and Assumptions}
We consider a stochastic, discrete-time, control-affine, nonlinear system subject to an additive Gaussian noise:
\[
s_{k+1}=f(s_k)+g(s_k)a_k+w_k,
\]
with state $s\in\R^n$, action $a\in\R^m$, internal dynamic $f:\mathbb{R}^n\to\mathbb{R}^n$, input-coupling matrix $g:\mathbb{R}^n\to\mathbb{R}^{n\times m}$, and $w$ the independent and identically distributed (i.i.d.), time-independent Gaussian noise process with zero mean and covariance matrix $\Sigma$.

For a stationary policy $\pi$, mapping states to actions through $\pi(s_k) = a_k$, the discounted infinite-horizon cost-to-go is 
\[
V^\pi(s)=\mathbb{E}\!\left[\sum_{k=0}^{\infty}\gamma^k\,c(s_k,\pi(s_k))\mid s_0=s\right],\quad \gamma \in (0,1),
\]
where $\gamma$ is the discount factor. The function $c:\mathbb{R}^n \times \mathbb{R}^m \to \mathbb{R}_{+}$ is the stage cost, separable into state (positive) and control (positive, strictly convex) components $q:\mathbb{R}^{n} \to \mathbb{R}$ and $\rho: \mathbb{R}^{m} \to \mathbb{R}$, such that $c(s,a)=q(s)+\rho(a)$. 
The optimal value $V^*$ satisfies the stochastic Bellman equation
\begin{equation}\label{eq:bellman}
    V^*(s) = \min_{a \in \mathbb{R}^m} \left\{ q(s) + \rho(a) + \gamma \mathbb{E}\big[V^*(f(s) + g(s)a + w)\big] \right\}
\end{equation}
and the optimal action $a^*$ is defined stepwise by:
\[
a^*_k = \arg \min_{a \in \mathbb{R}^m} \Big\{q(s_k)+\rho(a_k)+V^*(s_{k+1})\Big\},
\]
satisfying also the first-order optimality condition:
\begin{align}\label{eq:fooc1}
\nabla_a \rho(a)\Big|_{a=a^*}
+\gamma\,\nabla_a\mathbb{E}\!\left[V^*(f(s)+g(s)a+w)\right]\Big|_{a=a^*} = 0.
\end{align}

The stochastic converse optimality problem can eventually be summarized as follows. Given a chosen optimal value function $V^*: \mathbb{R}^n \to \mathbb{R}_+$, an input-coupling function $g$, a stage cost function $c(s,a) = q(s) + \rho(a)$, a discount factor $\gamma \in (0, 1)$, and a noise distribution $w \sim \mathcal{N}(0, \Sigma)$, find a function $f : \mathbb{R}^n \to \mathbb{R}^n$ such that the stochastic Bellman equation \eqref{eq:bellman} holds for all $s \in \mathbb{R}^n$. 

To guarantee that the stochastic Bellman equation is well-posed and that the first-order optimality condition is valid, we make the following set of assumptions. These are extensions of standard assumptions in stochastic optimal control to the converse optimality setting.\newline

\textbf{Assumption 1 (Well-posedness and regularity).}\label{ass:wellposed}
The following holds on $\mathbb{R}^n$:
\begin{enumerate}[leftmargin=1.35em,itemsep=2pt]
\item $q:\mathbb{R}^n\to\mathbb{R}_+$ is continuous.
\item $\rho:\mathbb{R}^m\to\mathbb{R}_+$ is $C^2$ and $\mu$-strongly convex: $\nabla^2\rho(a)\succeq \mu I$ for all $a$.
\item $f,g$ are locally Lipschitz with at most polynomial growth.
\item The candidate value function $V:\mathbb{R}^n\to\mathbb{R}$ is $C^2$.
\end{enumerate}

\textbf{Assumption 2 (Growth and integrability).}\label{ass:integrability}
There exist constants $\eta>0$ and an integer $\ell\ge 1$ such that for all $s\in\mathbb{R}^n$,
\[
|V(s)|+\|\nabla V(s)\|+\|\nabla^2 V(s)\|\le \eta\big(1+\|s\|^\ell\big),
\]
and, for each fixed $(s,a)\in\mathbb{R}^n\times\mathbb{R}^m$, the expectations
$\mathbb{E}[\|\nabla V(f(s)+g(s)a+w)\|]$ and $\mathbb{E}[\|\nabla^2 V(f(s)+g(s)a+w)\|]$ are finite under $w\sim\mathcal{N}(0,\Sigma)$.

\section{Stochastic Converse Optimality}
We now state the general converse result in the discounted, stochastic setting; a key technical step is the lemma justifying the interchange of expectation and differentiation in the first-order optimality condition.\newline

\textbf{Lemma 1.}\label{lem:interchange}
Under Assumptions 1 and 2, for all $s \in \mathbb{R}^n$ and $a \in \mathbb{R}^m$,
\begin{equation}
\begin{aligned}
\nabla_a\mathbb{E}\!\left[V(f(s)+g(s)a+w)\right]=\\
=\mathbb{E}\!\left[g(s)^\top\nabla V(f(s)+g(s)a+w)\right].
\end{aligned}
\end{equation}

\textit{Proof.} Fix $s \in \mathbb{R}^n$ and define $\phi(a,w) = V(f(s) + g(s)a + w)$. By the chain rule, we have
\[
\nabla_a \phi(a,w) = g(s)^{\top} \nabla V(f(s) + g(s)a + w)
\]
To interchange the derivative and the expectation, we verify the conditions of the dominated convergence theorem. Let $e_1,\dots, e_m$ be the canonical basis vectors of $\mathbb{R}^m$. 
Without loss of generality, we choose a sequence $\{t_{\nu}\}$ converging to $0$, such that we can define sequences $\{a_\nu^{(i)}\}$ through $a^{(i)}_{\nu} = a + t_{\nu}e_i$ converging to $a$ along each coordinate axis $i$.
By the mean value theorem, for each axis $i$, there exists $\xi^{(i)}_{\nu}$ on the line segment between $a^{(i)}_{\nu}$ and $a$ such that 
\begin{gather*}
\phi(a^{(i)}_{\nu}, w) - \phi(a,w) = \nabla_a \phi(\xi^{(i)}_{\nu},w)\cdot(a_{\nu}^{(i)} -a),\\ \text{and }
 |\phi(a^{(i)}_{\nu},w) - \phi(a,w)| \leq \|\nabla_a \phi(\xi^{(i)}_{\nu},w)\| \cdot \|a^{(i)}_{\nu} - a\|.
\end{gather*}

Now, let us bound $\|\nabla_a \phi(\xi^{(i)}_{\nu},w)\|$ for each axis $i$. From Assumption 2, we obtain:
\[
    \|\nabla_a \phi(\xi^{(i)}_{\nu},w)\| \leq \|g(s)\| \cdot \|\nabla V(f(s) + g(s)\xi^{(i)}_{\nu} + w)\|,
\]
and hence,
\[
\|\nabla_a \phi(\xi^{(i)}_{\nu},w)\| \le \|g(s)\| \cdot \eta(1 + \|f(s) + g(s)\xi^{(i)}_{\nu} + w\|^{\ell}).
\]

Since $\xi^{(i)}_{\nu}$ lies in a bounded neighborhood of $a$ (for $\nu$ sufficiently large), and $f,g$ have at most polynomial growth (Assumption 1.3), there exist constants $C_1(s)$ and $C_2(s)$ such that
\begin{align*}
\|f(s) + g(s)\xi^{(i)}_{\nu} + w\| 
&\leq C_1(s) + C_2(s)\|\xi^{(i)}_{\nu}\| + \|w\| \\
&\leq C(s,a) + \|w\|,
\end{align*}
where $C(s,a)$ is independent of $\nu$ and $\xi_{\nu}^{(i)}$ (i.e., uniform over the neighborhood of $a$). This uniformity ensures the subsequent growth bound 
\begin{align*}
\|\nabla_a \phi(\xi^{(i)}_{\nu},w)\| 
&\leq \eta\|g(s)\|(1 + (C(s,a) + \|w\|)^{\ell})\\
&\leq K(s,a)(1 +\|w\|^{\ell}),
\end{align*} 
where $K(s,a)(1 +\|w\|^{\ell})$ is independent of $\nu$. This provides a single bound applicable to the entire sequence. 
The growth bound $\beta :w \mapsto K(s,a)(1+\|w\|^l)$ is measurable. 

Let $\mu_\Sigma$ be the Gaussian measure. By definition of the expectation, the Lebesgue integral of $\beta$ weighted by the Gaussian density is 
\[\int_{\mathbb{R}^n}\beta(w) d\mu_{\Sigma}(w)=\mathbb{E}[K(s,a)(1+\|w\|^l)].\] 
The linearity of the expectation yields 
\[
\mathbb{E}[K(s,a)(1+\|w\|^l)] = K(s,a)(1+\mathbb{E}[\|w\|^l]),
\]
and hence,
\[\int_{\mathbb{R}^n}\beta(w) d\mu_{\Sigma}(w) = K(s,a)
\left(
1+\int_{\mathbb{R}^n}\|w\|^{\ell} d\mu_{\Sigma}(w)
\right).
\]

As all Gaussian moments are finite, we have:
\[
\int_{\mathbb{R}^n}\|w\|^{\ell} d\mu_{\Sigma}(w) = \mathbb{E}[\|w\|^{\ell}] < \infty
\]
for all $\ell>0$.
Hence, 
\[
\int_{\mathbb{R}^n}\beta(w) d\mu_{\Sigma}(w) < \infty,
\]
and the growth bound $\beta$ is Lebesgue-integrable with respect to the Gaussian measure. Since $\xi^{(i)}_{\nu} \to a$, continuity gives $\nabla_a \phi(\xi^{(i)}_{\nu},w) \to \nabla_a \phi(a,w)$ pointwise. Hence, the bound $\|\nabla_a \phi(\xi^{(i)}_{\nu},w)\| \le K(s,a)(1+\|w\|^\ell)$ provides a suitable dominating function, and by the dominated convergence theorem, we eventually obtain
\[
\lim_{\nu\to\infty} \mathbb{E}[\nabla_a \phi(\xi^{(i)}_{\nu},w)] = \mathbb{E}[\nabla_a \phi(a,w)].\quad (*)
\]

On the other hand, by definition of the gradient, we have 
\[
\nabla_a \mathbb{E}[\phi(a, w)] = \Big(\frac{\partial}{\partial a_1}\mathbb{E}[\phi(a, w)],\dots,\frac{\partial}{\partial a_m}\mathbb{E}[\phi(a, w)]\Big).
\]
The directional derivative along the $i$-th axis is 
\[
\frac{\partial}{\partial a_i}\mathbb{E}[\phi(a,w)] = \lim_{\nu\to\infty} \mathbb{E}\Big[\frac{\phi(a+t_{\nu}e_i,w) - \phi(a,w)}{t_{\nu}}\Big].
\]
Along the $i$-th axis, we also have 
\[
\frac{\phi(a+t_{\nu}e_i,w) - \phi(a,w)}{t_{\nu}} = \nabla_a \phi(\xi^{(i)}_{\nu}, w)\cdot e_i.
\] 
Combining these expressions yields 
\begin{gather*}
\lim_{\nu\to\infty} \mathbb{E}\Big[\frac{\phi(a+t_{\nu}e_i,w) - \phi(a,w)}{t_{\nu}}\Big] =\\= \lim_{\nu\to\infty} \mathbb{E}[\nabla_a \phi(\xi^{(i)}_{\nu}, w)\cdot e_i] = \mathbb{E}[\nabla_a \phi(a,w)\cdot e_i],
\end{gather*}
where the last equality follows from the dominated convergence argument $(*)$ above. But, $\nabla_a \phi(a, w) \cdot e_i = \frac{\partial}{\partial a_i}\phi(a, w)$. Therefore, $\frac{\partial}{\partial a_i}\mathbb{E}[\phi(a,w)] = \mathbb{E}[\frac{\partial}{\partial a_i}\phi(a, w)]$. Finally, we get 
\begin{gather*}
    \nabla_a \mathbb{E}[\phi(a,w)] = \Big(\mathbb{E}[\frac{\partial}{\partial a_1}\phi(a,w)], \dots, \mathbb{E}[\frac{\partial}{\partial a_m}\phi(a,w)]\Big) =\\= \mathbb{E}\Big(\frac{\partial \phi}{\partial a_1}(a,w), \dots,\frac{\partial \phi}{\partial a_m}(a,w)\Big)= \mathbb{E}[\nabla_a \phi(a,w)].
\end{gather*}\qed
\newline

\textbf{Theorem 1 (Stochastic converse optimality).}\label{thm:main}
Fix a stage cost $c(s,a) = q(s) + \rho(a)$ with $q,\rho$ satisfying Assumption 1, discount factor $\gamma \in (0,1)$, noise covariance $\Sigma \succ 0$, input coupling $g: \mathbb{R}^n \to \mathbb{R}^{n\times m}$, and a candidate value function $V^* \in C^2(\mathbb{R}^n)$ satisfying Assumption 2. Then there exists a drift $f: \mathbb{R}^n \to \mathbb{R}^n$ (i.e. a system $s_{k+1} = f(s_k) + g(s_k)a_k + w_k$ with $w_k \sim \mathcal{N}(0,\Sigma)$) for which $V^*$ is the optimal value function if and only if there exists a policy $\pi^*: \mathbb{R}^n \to \mathbb{R}^m$ such that for every $s \in \mathbb{R}^n$:
\begin{gather}
\nabla_a\Big[\rho(a)
+\gamma\,\mathbb{E}\!\left[V^*\big(f(s)+g(s)a+w\big)\right]\Big]\Big|_{a=a^*}=0,\label{eq:fooc}\\
V^*(s)=q(s)+\rho(a^*)+\gamma\mathbb{E}\!\left[V^*\big(f(s)+g(s)a^*+w\big)\right].\label{eq:bellman_id}
\end{gather}
When such $\pi^*$ exists, it is an optimal policy and $V^*$ satisfies the stochastic Bellman equation \eqref{eq:bellman} for all $s \in \mathbb{R}^n$.

\textit{Proof.} \textit{Sufficiency.} Suppose there exist a policy $\pi^*: \mathbb{R}^n \to \mathbb{R}^m$ and a drift $f: \mathbb{R}^n \to \mathbb{R}^n$ such that for every $s \in \mathbb{R}^n$, equations \eqref{eq:fooc} and \eqref{eq:bellman_id} hold, and $\pi^*(s_k) = a_k^*$ for all $s_k \in \mathbb{R}^n$. For a fixed $s\in \mathbb{R}^n$ we define the Bellman integrand:
\[
\Phi_s(a) := q(s) + \rho(a) + \gamma \mathbb{E}\big[V^*(f(s) + g(s)a + w)\big]
\]
By Lemma 1, we have:
\[
\nabla_a \Phi_s(a) = \nabla_a \rho(a) + \gamma \mathbb{E}\big[g(s)^\top \nabla V^*(f(s) + g(s)a + w)\big].
\] 
Equations (4)-(5) give $\nabla_a \Phi_s(a)|_{a=a^*} = 0$ and imply that the Bellman identity holds with action $a^*$. In particular, 
\begin{align*} 
V^*(s) &= q(s)+\rho(a^*)+\gamma \mathbb{E}\big[V^*(f(s)+g(s)a^*+w)\big]\\ 
&= \Phi_s(a^*). 
\end{align*} 
Since the Bellman equation involves the minimization of $\Phi_s(a)$ with respect to $a$, it follows that this minimum is attained at $a^*$. Hence: 
\begin{align*} 
\min_{a \in \mathbb{R}^m} \Phi_s(a) &= \Phi_s(a^*) \\
&= q(s) + \rho(a^*) + \gamma \mathbb{E}\big[V^*(f(s) + g(s)a^* + w)\big]\\ 
&= V^*(s). 
\end{align*} 
Thus, $\pi^*$ is an optimal policy and $V^*$ is the optimal value function for the drift $f$.

Now fix any stationary policy $\pi$ and define its Bellman operator:
\[
(T^\pi V)(s):=q(s)+\rho(\pi(s))+\gamma\mathbb{E}\!\left[V\big(f(s)+g(s)\pi(s)+w\big)\right].
\]
By definition of the minimum, $\min_a \Phi_s(a)\le \Phi_s(\pi(s))$, hence $V^*(s)\le (T^\pi V^*)(s)$ for all $s$. Applying $T^\pi$ to both sides of $V^* \le T^\pi V^*$ and using the monotonicity of $T^\pi$ 
(which follows from the positivity of costs), we obtain $T^\pi V^* \le (T^\pi)^2 V^*$. By induction, $V^* \le (T^\pi)^k V^*$ for all $k \in \mathbb{N}$. Since $T^\pi$ is a $\gamma$-contraction in an appropriate weighted norm (e.g. the sup norm under bounded costs, or a weighted norm under the discounted criterion), then by the Banach fixed-point theorem, $(T^\pi)^k V^* \to V^\pi$, the value function of policy $\pi$. Taking limits preserves the inequality, yielding $V^* \le V^\pi$ on $\mathbb{R}^n$, proving the optimality of $V^*$ and $\pi^*$.

\textit{Necessity.}
Assume $V^*$ optimal, so for each $s$ the minimizer $\pi^*(s)$ in \eqref{eq:bellman} exists. 
By the measurable selection theorem \cite{shreve1979universally}, $\pi^*$ is chosen to be measurable, and hence admissible as a feedback policy. At a minimum, we have $\nabla_a\Phi_s(\pi^*(s))=0$. Lemma 1 justifies interchanging the derivative and the expectation, yielding \eqref{eq:fooc}. The Bellman identity \eqref{eq:bellman_id} holds by the definition of $\pi^*(s) = a^*$ as the minimizer in \eqref{eq:bellman}. \qed

\section{Quadratic--Gaussian Specialization and Constructive Drifts}
We now examine the Quadratic--Gaussian (QG) setting, where the cost functions and the value function are now quadratic. In this setting, $a^*$ will be shown to be unique. With explicit $V^*$ and $a^*$, we can generate and validate numerous instances for RL evaluation. 
Let $b$ be the certainty equivalence offset. We assume the following:
\[
q(s)=s^\top Qs,\quad \rho(a)=a^\top Ra,\quad V^*(s)=s^\top Ps+b,
\]
with $Q\succeq 0$, $R\succ 0$, $P\succ 0$, and $w\sim\mathcal{N}(0,\Sigma)$.\newline

\textit{Certainty-equivalent optimal policy.} We start from the Bellman integrand:
\[
    \Phi_s(a) = q(s) + \rho(a)+\gamma\,\mathbb{E}[V^*(f(s)+g(s)a+w)]. 
\]
For ease of reading, let $z:=z(s,a) = f(s)+g(s)a$. As $V^*(s) = s^{\top}Ps + b$, we have:
\begin{gather*}
\mathbb{E}[V^*(z+w)] = \mathbb{E}[(z+w)^{\top}P(z+w) + b] =\\
=z^{\top}Pz + 2z^{\top}P\mathbb{E}[w] + \mathbb{E}[w^{\top}Pw] + b.
\end{gather*}
Since $w\sim\mathcal{N}(0, \Sigma)$, we have $\mathbb{E}[w] = 0$ and $\mathbb{E}[w^{\top}Pw] = \text{tr}(P\Sigma)$, and we are left with
\[
\mathbb{E}[V^*(z+w)] = z^{\top}Pz + \text{tr}(P\Sigma) + b.
\]
Hence, it follows that:
\[
\Phi_s(a) = s^{\top}Qs + a^\top R a +\gamma\,z^\top Pz+\gamma\,\mathrm{tr}(P\Sigma)+\gamma b,
\]
and
\[
\nabla_a \Phi_s(a) = \nabla_a\Big[a^{\top}Ra + \gamma z^{\top}Pz\Big],
\]
as the others terms do not depend on $a$. We then have:
\begin{align*}
\nabla_a \Phi_s(a) &= \nabla_a\Big[a^{\top}Ra + \gamma\Big(f(s)^{\top}Pz + a^{\top}g^{\top}Pz\Big)\Big]\\
&= \nabla_a\Big[a^{\top}Ra + \gamma \Big(f(s)^{\top}Pf(s) + f(s)^{\top}Pg(s)a +\\&\qquad\qquad\text{ }+ a^{\top}g(s)^{\top}Pf(s) + a^{\top}g(s)^{\top}Pg(s)a\Big)\Big]\\
&= \nabla_a\Big[a^{\top}B(s)a + 2\gamma a^{\top}g(s)^{\top}Pf(s) +\\
&\qquad\qquad\qquad\qquad\qquad\qquad\quad+\gamma f(s)^{\top}Pf(s)\Big],
\end{align*}
where $B(s)=R+\gamma g(s)^\top P g(s)$. Setting $\nabla_a\Phi_s(a) = 0$ yields:
\[
\nabla_a \Phi_s(a) = 2B(s)a + 2\gamma g(s)^{\top}Pf(s) = 0.
\]
Solving for $a$ leads to the stationary optimal feedback
\[
a^*=-\gamma\,B(s)^{-1}g(s)^\top P f(s).
\]
As $\nabla_a^2\Phi_s(a) = 2B(s)\succ0$, $\Phi_s$ is strictly convex and its minimizer $a^*$ is unique.

\textit{Discounted metric and energy identity.}
The minimized term $\min_{a\in\mathbb{R}^m}\Phi_s(a) = \min_{a\in\mathbb{R}^m} \Big[a^{\top}Ra + \gamma z^\top Pz\Big] = f(s)^{\top}H_\gamma(s) f(s)$ induces a state-dependent metric:
\begin{equation}\label{eq:Hgamma}
\begin{aligned}
H_\gamma(s)
&=\gamma\Big(P-\gamma\,P g(s)B(s)^{-1}g(s)^\top P\Big)\\
&=\gamma\Big(P^{-1}+\gamma\,g(s)R^{-1}g(s)^\top\Big)^{-1}\succ 0,
\end{aligned}
\end{equation}
where the second equality uses the Woodbury identity. Inserting $\min_{a\in\mathbb{R}^m}\Phi_s(a)$ into the Bellman equation~\eqref{eq:bellman} leads to
\begin{align*}
V^*(s) &= s^{\top}Ps + b\\
&= \min_{a \in \mathbb{R}^m} \left\{ q(s) + \rho(a) + \gamma \mathbb{E}\big[V^*(z + w)\big] \right\}\\
&= q(s) + \min_{a \in \mathbb{R}^m} \left\{a^{\top}Ra + \gamma z^{\top}Pz \right\} + \gamma \text{tr}(P\Sigma) + \gamma b\\
&= q(s) + f(s)^{\top}H_{\gamma}(s)f(s) + \gamma \text{tr}(P\Sigma) + \gamma b.
\end{align*}
Comparing the state-dependent parts gives
\[
s^{\top}Ps = q(s) + f(s)^{\top}H_{\gamma}(s)f(s),
\]
and reduces the Bellman identity to the quadratic ``energy'' equation: 
\[
s^\top(P-Q)s = f(s)^\top H_\gamma(s)\,f(s).
\]
In the generator, this energy equation 
is the design equation: it characterizes the family of drifts that make the prescribed $V^*$ optimal.

\textit{Noise offset.}
Comparing the constant parts gives
\[
    b = \gamma \text{tr}(P\Sigma) + \gamma b,
\]
and hence additive Gaussian noise contributes only a constant shift in value, with
\[
b = \frac{\gamma}{1-\gamma}\text{tr}(P\Sigma).
\]

\textit{Two constructive parameterizations.}
All drifts satisfying the energy equation 
admit explicit equivalent square-root (SR) and metric-normalized (MN) parameterizations that make it easy to inject bounded nonlinear structure.
For any nonzero direction field $\tilde f(s)$, the square-root parameterization is given by:
\[
f(s)=\sqrt{s^\top(P-Q)s}\;\frac{H_\gamma(s)^{-1/2}\tilde f(s)}{\|\tilde f(s)\|}, \tilde f(s)\neq 0.
\]
Equivalently, for any orthogonal field $S(s)$ with $S(s)^\top S(s)=I$,
\[
f(s)=H_\gamma(s)^{-1/2}S(s)(P-Q)^{1/2}s.
\]

Our benchmarks primarily use 
this latter equation: orthogonality guarantees 
the energy equation 
exactly, while the field $S(\cdot)$ provides a knob for nonlinear coupling and stiffness.

\section{Benchmark Generation, Families, and Dataset}
  
We now describe the systematic construction of benchmark instances based on the Quadratic-Gaussian (QG) specialization developed in Section V. The goal is to generate a diverse set of stochastic control problems, each with a certified optimal policy $\pi^*$ and an optimal value function $V^*$, which can be used to evaluate reinforcement learning algorithms. We first define the general problem class and the generator pipeline. We then introduce two benchmark families: (i) a nonlinear linked-arm system with state-dependent actuation and (ii) a dynamically-extended nonholonomic vehicle model. Finally, we describe how validated instances are assembled into a dataset for algorithm evaluation.

\subsection{Problem definition and generation of benchmark families}

Each benchmark instance is a discrete-time stochastic control problem of the form:
\[
s_{k+1} = f(s_k) + g(s_k)a_k + w_k,
\]
with state $s\in\R^n$, action $a\in\R^m$, internal dynamic $f:\mathbb{R}^n\to\mathbb{R}^n$, input-coupling matrix $g:\mathbb{R}^n\to\mathbb{R}^{n\times m}$, a discount factor $\gamma \in (0,1)$, and $w$ the independent and identically distributed (i.i.d.), time-independent Gaussian noise process with zero mean and covariance matrix $\Sigma$. The corresponding stage cost $c:\mathbb{R}^n \times \mathbb{R}^m \to \mathbb{R}$ is defined by $c(s,a) = q(s) + \rho(a)$. The functions $f$, $g$, $q$, and $\rho$ satisfy Assumptions 1 and 2. The optimal policy $\pi^*:\mathbb{R}^{n}\to\mathbb{R}^{m}$ and the optimal value function $V^*:\mathbb{R}^{n}\to\mathbb{R}$ are analytically known from the converse-optimal construction of Section V.

A \emph{benchmark family} is a parametric collection of control problems:
\[
\mathcal{F}_\psi = \{ M_\psi : \psi \in \Psi \},
\]
where each control problem $M_\psi$ is defined by a tuple $(f_\psi, g_\psi, q_\psi, \rho_\psi, \Sigma_\psi)$ that satisfies the QG conditions of Section V, and $\Psi$ is a parameter space. The certified ground truth $(\pi^*_\psi, V^*_\psi)$ follows directly from the construction.

We construct two distinct families, $\mathcal{F}_{\text{Arm}}$ and $\mathcal{F}_{\text{NVDEx}}$, which are described below. Each family provides a range of difficulty levels through interpretable parameters; we make the physical structure explicit through governing equations. Depending on the structure of the input matrix, two equivalent QG parameterizations are convenient: a state-dependent formulation used for the Arm family, a constant-input formulation used for the NVDEx family. In addition, the converse-optimal drift is not treated as an abstract matrix field only: its factors are chosen so that they admit a robotics interpretation. In particular, the orthogonal matrix field $S(\cdot)$ is implemented as a product of planar rotations, which preserves the converse-optimal energy identity exactly while producing dynamics that resemble physically meaningful transition and coupling operators in robotic systems. The generator supports multiple families; here we focus on two physically grounded, scalable constructions that stress different failure modes of deep RL.

The generator (Algorithm~\ref{alg:generator}) produces instances with a validation trace. Each instance undergoes three checks:
\begin{enumerate}
\item \textbf{Symmetric positive-definiteness (SPD) of $B(s)$}: 
\[B(s)=R+\gamma g(s)^\top P g(s)\succ 0, s \in \mathbb{R}^n.\]
The SPD of $B(s)$ is required for two reasons: (i) it guarantees strict convexity of the Bellman integrand $\Phi_s(a)$, so the first-order condition yields a unique global minimizer $a^*$; and (ii) it keeps the discounted metric $H_\gamma(s)$ positive definite, which is necessary for the energy equation $s^\top(P-Q)s = f(s)^\top H_\gamma(s)f(s)$ to be well-posed and numerically stable. In practice, numerically verifying $B(s)\succ 0$ on a sampling grid over a prescribed bounded test region in the domain $\mathbb{R}^n$ guarantees the mathematical consistency of the constructed benchmark instance.
\item \textbf{Bellman residual}: this check verifies that the constructed $(f,g,q,\rho)$ indeed satisfy the optimality conditions numerically. For each test point $s$ on a grid $\mathcal{G}$, we compute:
\begin{align*}
    \delta(s) &= \Big|V^*(s) - \Big(q(s)+\rho(a^*(s)) +\\ &\qquad\qquad\quad+ \gamma\mathbb{E}[V^*(f(s)+g(s)a^*(s)+w)]\Big)\Big|.
\end{align*}
If $\max_{s\in\mathcal{G}} \delta(s) < \varepsilon$, where $\varepsilon$ is a suitable numerical tolerance threshold, the instance passes validation. This avoids implementation errors and ensures that the benchmark is correctly instantiated.
\item \textbf{Boundedness}: the boundedness check simulates the closed-loop system under the optimal policy $\pi^*$ over a fixed horizon and domain. This verifies that numerical implementation does not introduce instabilities, that the theoretically optimal behavior translates into practice, and that the benchmark instance is usable for RL evaluation, ensuring trajectories remain well-behaved and do not diverge due to accumulated errors or hidden sensitivities. Instances that fail this test are rejected, guaranteeing that every exported fixture is both mathematically correct and practically stable.
\end{enumerate}
Validated instances are exported as YAML fixtures containing parameters, CRN seeds, and reference evaluators for $V^*$ and $a^*$.

\begin{algorithm}[h]
\caption{Converse-Optimal Benchmark Generator (QG case)}
\label{alg:generator}
\footnotesize
\begin{algorithmic}[1]
\STATE \textbf{Input:} family type, parameter ranges, number of instances $N$, parameterization (\texttt{sqrt} or \texttt{metric}), nonlinearity knob $\alpha$, seeds
\FOR{each hyperparameter tuple $\psi$}
  \STATE Sample parameter vector $\psi_i$ from family-specific distribution $P_\Psi$
  \STATE Extract $(n, m, Q, R, P, \gamma, \Sigma)$ and input map $g$ from $\psi_i$
  \STATE Verify $R+\gamma g(s)^\top P g(s)\succ 0$ on domain
  $\mathcal D$
  \STATE Choose direction field $\tilde f(s)$ or orthogonal field $S(s)$ (bounded via $\alpha$)
  \STATE Construct $f$ via parameterizations; compute $a^*(s)$
  \STATE Validate Bellman identity on a grid; test closed-loop boundedness under $a^*$
  \STATE Export fixture (YAML) with reference $(a^*,V^*)$ and CRN seeds
\ENDFOR
\STATE \textbf{Output:} fixture dataset + validator + reference solvers + plotting scripts
\end{algorithmic}
\end{algorithm}

\subsection{Serial $n$-link planar arm systems (\texttt{ConverseArm})}

The first family, $\mathcal{F}_{\text{Arm}}$, models a serial $n$-link planar arm with revolute joints.\newline

\subsubsection{State and action spaces}

In order to do this, we must express the configuration (joint angles) and the rates of change (angular velocities) of each joint. The natural choice is therefore to include for each joint $i$:
\begin{itemize}
\item $\theta_i$ — the angular position (angle) of the joint $i$;
\item $\omega_i$ — the angular velocity of the joint $i$.
\end{itemize}
Stacking these variables for all $n$ joints gives the state vector
\[
s = [\theta_1, \omega_1, \theta_2, \omega_2, \ldots, \theta_n, \omega_n]^\top \in \mathbb{R}^{2n},
\]
In this ordering, each joint occupies two consecutive components: position followed by velocity. Thus, for joint $i$, the angle $\theta_i$ appears at position $2i-1$ (odd index), and the angular velocity $\omega_i$ appears at position $2i$ (even index). This convention is convenient when writing the dynamics and the input coupling matrix, as it allows us to refer to the velocity component of the joint $i$ simply as the $2i$-th entry of the state vector.

The control inputs are the torques applied at each joint:
\[
a = [\tau_1, \tau_2, \ldots, \tau_n]^\top \in \mathbb{R}^n,
\]
where $\tau_i$ is the torque commanded at the joint $i$. Positive torques act to increase the corresponding angular velocity. 

This choice of state and action spaces is standard for robotic manipulators: the state contains the information needed to predict future evolution (positions and velocities), while the actions represent the physical inputs available to the controller (joint torques). The dimension grows linearly with the number of joints, allowing systematic scaling to higher-dimensional problems.\newline

\subsubsection{Dynamics}

The discrete-time dynamics are obtained by Euler discretization of the continuous-time equations of motion, with sampling step $\tau>0$. The kinematic update for each joint angle is straightforward:
\[
\theta_{i,k+1} = \theta_{i,k} + \tau\,\omega_{i,k} + w^{\theta}_{i,k},
\]
where $w^{\theta}_{i,k}$ is the component of the process noise affecting the angle of the joint $i$.

The update in angular velocity is more complex because it depends on both the internal dynamics (torques arising from motion, gravity, and coupling between joints) and the applied control torques. In our converse-optimal construction, the internal dynamics are encoded in the drift $f_p(s)$, while the control enters through the input matrix $g(s)$. Specifically, the angular velocity update is:
\[
\omega_{i,k+1} = [f_p(s_k)]_{2i} + p\,\cos(\theta_{i,k})\,a_{i,k} + w^{\omega}_{i,k},
\]
where:
\begin{itemize}
\item $[f_p(s_k)]_{2i}$ denotes the $2i$-th component of the drift vector (recall the state ordering $[\theta_1,\omega_1,\ldots,\theta_n,\omega_n]$, so odd indices correspond to angles, even indices to angular velocities);
\item $p\in(0,1]$ is a control-strength continuation parameter that scales the input. When $p=1$, the system has full control authority; as $p\to 0$, control becomes weaker;
\item $\cos(\theta_{i,k})$ captures the configuration-dependent mechanical advantage: torque applied at joint $i$ is most effective when the joint angle is near zero (torque acts perpendicular to the link), and becomes ineffective as $\theta_i\to\pm\frac{\pi}{2}$ (torque aligns with the link direction);
\item $w^{\omega}_{i,k}$ is the process noise affecting the angular velocity of joint $i$.
\end{itemize}

The complete system can be written in the compact control-affine form:
\[
s_{k+1} = f_p(s_k) + p\,g(s_k)a_k + w_k,\quad w_k\sim\mathcal{N}(0,\Sigma),
\]
where the input matrix $g(s)$ has entries
\[
[g(s)]_{2i,i} = \cos\theta_i,\quad i=1,\ldots,n,
\]
with all other entries zero. This structure ensures that each torque $a_i$ affects only its corresponding angular velocity channel, scaled by the mechanical advantage $\cos\theta_i$, exactly as in the governing equations above.\newline

\subsubsection{Drift construction}

The drift $f_p$ is not specified \textit{a priori}; it is constructed using the converse-optimality framework to ensure that the prescribed value function $V^*(s)=s^\top P s + b$ is optimal. Using the MN parameterization, we set
\[
f_p(s) = H_\gamma^{(p)}(s)^{-1/2} S_p(s) (P-Q)^{1/2} s,
\]
where $H_\gamma^{(p)}(s)$ is the discounted metric with control strength $p$:
\[
H_\gamma^{(p)}(s) = \gamma\big(P - \gamma p^2 P g(s)(R + \gamma p^2 g(s)^\top P g(s))^{-1} g(s)^\top P\big).
\]

The orthogonal field $S_p(s)$ is implemented as a product of Givens rotations:
\begin{align*}
S_p(s) &= \left[\prod_{i=1}^{n-1} G(\omega_i,\omega_{i+1}; \beta_i(s,p))\right] \times \\ &\qquad\qquad\qquad\qquad\quad\times\left[\prod_{i=1}^{n} G(\theta_i,\omega_i; \alpha_i(s,p))\right],
\end{align*}
with rotation angles
\begin{align*}
\alpha_i(s,p) &= \alpha_0 p \tanh(\kappa_1 \theta_i \omega_i) + g_0 p \sin\theta_i,\\
\beta_i(s,p) &= \beta_0 p \tanh(\kappa_2(\theta_{i+1}-\theta_i)).
\end{align*}
These angles introduce physically interpretable couplings:
\begin{itemize}
\item The $\tanh(\kappa_1 \theta_i \omega_i)$ term creates a velocity-dependent coupling similar to Coriolis forces in robotic manipulators;
\item The $\sin\theta_i$ term introduces a gravity-like bias that depends on joint angle. $g_0$ scales the magnitude of this bias term;
\item The $\tanh(\kappa_2(\theta_{i+1}-\theta_i))$ term couples neighboring joints, similar to elastic or damping forces in a serial linkage.
\end{itemize}
The constants $\kappa_1$ and $\kappa_2$ are gain parameters that control the sensitivity of the coupling terms:
\begin{itemize}
\item $\kappa_1$ scales the argument of the hyperbolic tangent in $\alpha_i$, determining how strongly the product $\theta_i\omega_i$ (position–velocity coupling) influences the rotation angle. Larger $\kappa_1$ makes the $\tanh$ saturate more quickly, meaning the coupling becomes active even for small deviations from zero;
\item $\kappa_2$ scales the argument of the hyperbolic tangent in $\beta_i$, controlling the sensitivity of the neighbor-coupling term to the angle difference $\theta_{i+1}-\theta_i$. Higher $\kappa_2$ makes the coupling more responsive to small misalignments between adjacent joints.
\end{itemize}

In both cases, the hyperbolic tangent ensures that the coupling remains bounded: $|\tanh(\cdot)| < 1$, so the rotation angles $\alpha_i$ and $\beta_i$ are confined to intervals determined by $\alpha_0 p$ and $\beta_0 p$ respectively. This boundedness guarantees that the orthogonal field $S_p(s)$ remains well-defined and numerically stable across the entire state space. Crucially, because each $G$ is an orthogonal rotation matrix, the product $S_p(s)$ remains orthogonal ($S_p^\top S_p = I$). This orthogonality guarantees that the energy equation $s^\top(P-Q)s = f_p(s)^\top H_\gamma^{(p)}(s) f_p(s)$ holds exactly, preserving the converse-optimal construction while injecting nonlinear, physically motivated coupling.\newline

\subsubsection{Optimal policy and value}

With the drift constructed as above, the optimal action derived from the QG specialization is:
\[
a_p^* = -\gamma p\big(R + \gamma p^2 g(s)^\top P g(s)\big)^{-1} g(s)^\top P f_p(s).
\]
The parameter $p\in (0,1]$ scales $g\mapsto pg$, resulting in a homotopy family that preserves the analytic structure of the optimal action $a_p^*$. The optimal policy provides the control authority that, together with the drift $f_p(s)$, steers the system optimally according to $V^*$.\newline

\subsubsection{Difficulty knobs}

The family is parameterized by 
\[
\psi = (n, p, Q, R, P, \Sigma, \alpha_0, \beta_0, \kappa_1, \kappa_2, g_0).
\]
To facilitate ordering benchmark instances by expected difficulty, we define a scalar heuristic index:
\[
D = \lambda_1 n + \lambda_2 p^{-2} + \lambda_3 \text{tr}(\Sigma) + \lambda_4 \alpha_0 + \lambda_5 \kappa(P),
\]
with fixed weights $\lambda_i>0$. This index combines:
\begin{itemize}
\item $n$: dimensionality (more joints = more complex dynamics);
\item $p^{-2}$: penalizes weak control authority (small $p$), as it makes stabilization harder;
\item $\tr(\Sigma)$: noise power;
\item $\alpha_0$: nonlinearity gain of $S_p$ (larger $\alpha_0$ introduces stronger coupling);
\item $\kappa(P)$: condition number of $P$, a proxy for metric ill-conditioning;
\end{itemize}
Although $D$ is heuristic, it provides a useful relative ordering of instances for ablation studies and difficulty sweeps.\newline

While the \texttt{ConverseArm} family emphasizes serial-chain dynamics with state-dependent actuation, the next family takes a different structural form. The \texttt{NVDEx} system features constant input matrices and includes a controlled open-loop instability mechanism, offering complementary challenges to reinforcement learning algorithms.

\begin{figure}
\centering
\incfig[width=0.95\linewidth]{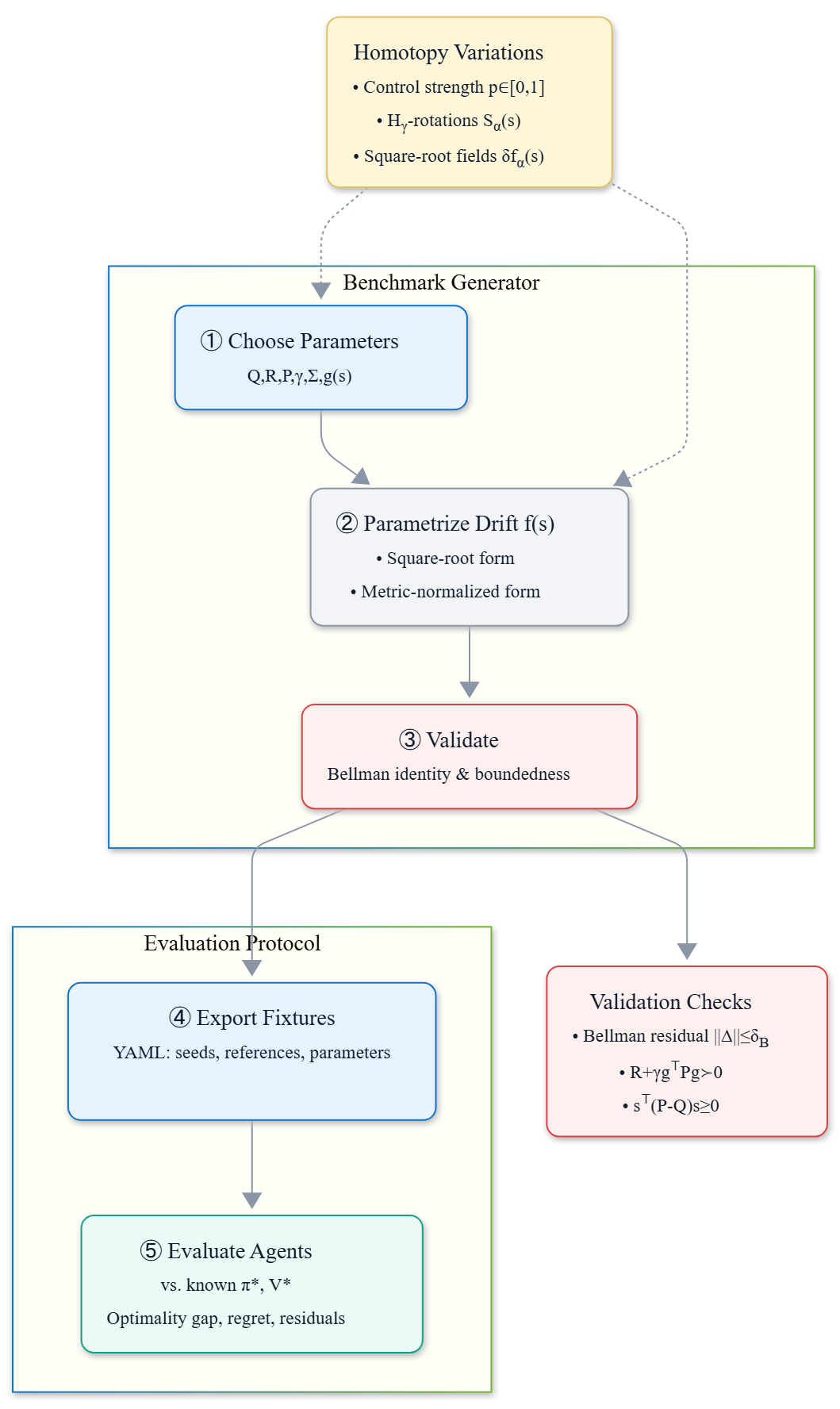}
\caption{Converse-optimal benchmark pipeline: sampling, drift construction, validation, fixture export, paired evaluation, and reporting.}
\label{fig:pipeline}
\end{figure}

\subsection{Nonholonomic Vehicle with Dynamic Extension (NVDEx)}

The second family, $\mathcal{F}_{\text{NVDEx}}$, models a unicycle/differential-drive vehicle module with dynamic extension, allowing unstable open-loop configurations.\newline

\subsubsection{State and action spaces}

A standard unicycle model has states $(x,y,\phi)$ representing position and heading, with controls for linear velocity $v$ (magnitude of velocity in direction of heading) and angular velocity $\omega$ (rate of change of heading). However, such a model is too simple for challenging benchmarks: control inputs directly set velocities, leaving little room for complex dynamics.

We therefore introduce the dynamic extension: instead of commanding velocities directly, we command their derivatives (accelerations), and the velocities themselves become part of the state. This creates a richer dynamical structure where:
\begin{itemize}
\item The system has inertia: the velocity changes only through applied accelerations;
\item Multiple modules $K$ can be coupled, increasing the dimensionality.
\end{itemize}

For $j > 0$, we denote $v_i^{(j)}$ and $\omega_i^{(j)}$ the internal states in the dynamic extension chain and the angular velocity chain, respectively. They represent ``delayed'' or ``filtered'' versions of the velocity, creating inertia-like effects. The integrator depths $r_v$ and $r_\omega$ control how many steps of ``memory'' appear between the acceleration commands and the actual velocity changes. Integrator depth refers to how many cascaded integrators lie between the control input and the actual physical velocity. For $r_v = 1$, the chain contains only the linear velocity $v_i^{(0)}$, which is directly affected by the control; the response is immediate, with low inertia. For $r_v = 2$, the chain contains $v_i^{(0)}$ and $v_i^{(1)}$; control affects $v_i^{(1)}$, which then integrates to influence $v_i^{(0)}$, creating a smoother and more realistic response with inertia. For $r_v = 3$, the chain has three levels, further increasing the smoothing between the command and the actual velocity. Higher integrator depth makes the system more challenging to control because the effect of control actions is delayed and filtered, the state space dimension increases, and the dynamics become more complex, requiring the controller to anticipate future effects of current actions. Thus, $r_v$ and $r_\omega$ serve as difficulty knobs that increase dynamical complexity while preserving the certified optimal policy structure.

For a system with $K\geq 1$ modules and integrator depths $r_v, r_\omega \in \mathbb{N}$, the state vector of module $i$ contains:
\begin{itemize}
\item Position $(x_i, y_i)$ and heading $\phi_i$ (the core unicycle states);
\item A chain of $r_v$ velocity variables $v_i^{(0)},\ldots,v_i^{(r_v-1)}$, where $v_i^{(0)}$ is the actual linear velocity, and the higher indices represent intermediate states in the extension;
\item A chain of $r_\omega$ angular velocity variables $\omega_i^{(0)},\ldots,\omega_i^{(r_\omega-1)}$, similarly structured.
\end{itemize}
Thus, the state $s_i \in \mathbb{R}^{3 + r_v + r_\omega}$ of the module $i$ is:
\[
s_i = [x_i, y_i, \phi_i, v_i^{(0)},\ldots,v_i^{(r_v-1)}, \omega_i^{(0)},\ldots,\omega_i^{(r_\omega-1)}]^\top.
\]
The action for the module $i$ consists of the accelerations commanded to the velocity chains:
\[
a_i = [u_{v,i}, u_{\omega,i}]^\top \in \mathbb{R}^2,
\]
where $u_{v,i}$ accelerates the linear velocity chain and $u_{\omega,i}$ accelerates the angular velocity chain.

For multi-module systems ($K>1$), states and actions are simply stacked:
\begin{align*}
s &= [s_1^\top, s_2^\top, \ldots, s_K^\top]^\top \in \mathbb{R}^{(3+r_v+r_\omega)K},\\
a &= [a_1^\top, a_2^\top, \ldots, a_K^\top]^\top \in \mathbb{R}^{2K}.
\end{align*}

Analogously to the \texttt{ConverseArm} setup, this choice of state and action spaces reflects the structure of a unicycle with dynamic extension: the module's configuration consists of its position and heading; velocities determine its motion through kinematic equations; inertia by placing integrators between acceleration commands and actual velocities. Dimension scales linearly with the number of modules $K$, allowing systematic progression from single-module to multi-agent benchmarks while maintaining physical interpretability and certified optimality.\newline

\subsubsection{Dynamics}

With the sampling step $\tau>0$, the discrete-time dynamics reflects the physical relationships between these variables. For each module $i$, the position and heading evolve according to the actual linear and angular velocities, and are obtained by Euler discretization of continuous-time kinematic equations:
\begin{align*}
x_{i,k+1} &= x_{i,k} + \tau\,v_{i,k}^{(0)}\cos\phi_{i,k} + w^x_{i,k},\\
y_{i,k+1} &= y_{i,k} + \tau\,v_{i,k}^{(0)}\sin\phi_{i,k} + w^y_{i,k},\\
\phi_{i,k+1} &= \phi_{i,k} + \tau\,\omega_{i,k}^{(0)} + w^\phi_{i,k}.
\end{align*}

The velocity chains propagate their values upward: each level feeds into the next, like a cascade of integrators. For $j = 0, \ldots, r_v-2$:
\[
v_{i,k+1}^{(j)} = v_{i,k}^{(j)} + \tau\,v_{i,k}^{(j+1)} + w^{v,j}_{i,k},
\]
and similarly for $j = 0, \ldots, r_\omega-2$:
\[
\omega_{i,k+1}^{(j)} = \omega_{i,k}^{(j)} + \tau\,\omega_{i,k}^{(j+1)} + w^{\omega,j}_{i,k}.
\]

At the highest level of each velocity chain, the control input directly influences the dynamics. This is where the commanded acceleration enters the system. For the linear velocity chain, the top level is $v_i^{(r_v-1)}$. Its update equation is:
\[
v_{i,k+1}^{(r_v-1)} = [f(s_k)]_{v,i} + \tau\,u_{v,i,k} + w^{v,r_v-1}_{i,k}.
\]
where we have:
\begin{itemize}
\item $\tau\,u_{v,i,k}$: the direct effect of the control input. The sampling step $\tau$ scales the acceleration command $u_{v,i,k}$ to produce a change in this top-level state over one discrete time step;
\item $[f(s_k)]_{v,i}$: this term comes from the converse-optimal drift construction. It represents the contribution of the system's internal dynamics, i.e., how the top-level state would evolve even in the absence of control;
\item $w^{v,r_v-1}_{i,k}$: the process noise affecting this state, drawn from the Gaussian distribution with covariance $\Sigma$.
\end{itemize}
An analogous equation holds for the angular velocity chain, with $u_{\omega,i,k}$ the commanded angular acceleration:
\[
\omega_{i,k+1}^{(r_\omega-1)} = [f(s_k)]_{\omega,i} + \tau\,u_{\omega,i,k} + w^{\omega,r_\omega-1}_{i,k}.
\]
After stacking all modules, these equations collapse into the compact control-affine form used throughout the paper:
\[
s_{k+1} = f_p(s_k) + p G a_k + w_k,\quad w_k\sim\mathcal{N}(0,\Sigma),
\]
with control-strength continuation $p\in(0,1]$.\newline

\subsubsection{Input coupling}

Each control input $a_i = [u_{v,i}, u_{\omega,i}]^\top$ influences exactly one state variable (the top of its respective chain), and it does so linearly with coefficient $\tau$. This local, linear influence can be encoded in a matrix $G$ that maps the global action vector $a = [a_1^\top, \ldots, a_K^\top]^\top$ to the correct components of the state update. $G$ is a matrix that selects which states receive control inputs and places $\tau$ in those positions. This gives $G$ a block-diagonal structure:
\[
G = \mathrm{blkdiag}(\underbrace{G_{\text{mod}}, G_{\text{mod}}, \ldots, G_{\text{mod}}}_{K \text{ times}}) \in \mathbb{R}^{n \times 2K}.
\]

The per-module matrix $G_{\text{mod}}$ encodes how accelerations $u_{v,i}$ and $u_{\omega,i}$ enter the dynamics. The control inputs $u_{v,i}$ and $u_{\omega,i}$ should affect only the top-level states $v_i^{(r_v-1)}$ and $\omega_i^{(r_\omega-1)}$, respectively. Therefore, $G_{\text{mod}}$ is a matrix with $3+r_v+r_\omega$ rows and $2$ columns, where:
\begin{itemize}
\item All rows corresponding to $x_i, y_i, \phi_i$, and lower-level velocity states are zero; controls do not directly affect these.
\item The row corresponding to $v_i^{(r_v-1)}$ (the top of the linear velocity chain) has a $1$ in the first column (for $u_{v,i}$) and $0$ in the second.
\item The row corresponding to $\omega_i^{(r_\omega-1)}$ (the top of the angular velocity chain) has a $0$ in the first column and a $1$ in the second column (for $u_{\omega,i}$).
\end{itemize}
Explicitly, with the sampling step $\tau$ factored out, we have:
\[
G_{\text{mod}} = \tau\begin{bmatrix}
\mathbf{0}_{(3+r_v-1)\times 2} \\[2pt]
(1,\ 0) \\[2pt]
\mathbf{0}_{(r_\omega-1)\times 2} \\[2pt]
(0,\ 1)
\end{bmatrix} \in \mathbb{R}^{(3+r_v+r_\omega)\times 2}.
\]

\subsubsection{Drift construction}

Following the converse-optimal framework, the drift $f_p(s)$ is constructed using the metric-normalized parameterization:
\[
f_p(s) = (H_\gamma^{(p)})^{-1/2} S_\nu(s) (P-Q)^{1/2} s,
\]
where the discounted metric with control strength $p$ is:
\[
H_\gamma^{(p)} = \gamma\big(P^{-1} + \gamma p^2 G R^{-1} G^\top\big)^{-1} \succ 0.
\]
Note that unlike the \texttt{ConverseArm} family, $H_\gamma^{(p)}$ is state-independent because $G$ is constant.

Here, $S_\nu(s)$ is an orthogonal field which introduces nonlinear coupling while preserving the energy identity. It is constructed as a product of bounded rotations:
\begin{itemize}
\item A rotation in the $(x_i, y_i)$ plane by the local heading $\phi_i$;
\item A coupling between $(v_i^{(0)}$ and $\omega_i^{(0)})$ via angle $\nu_i(s) = \alpha \tanh(\kappa v_i^{(0)}\omega_i^{(0)})$, which creates an interaction between linear and angular velocities;
\item Optional cross-couplings (e.g., between $(x_i, v_i^{(0)})$ and $(y_i, \omega_i^{(0)})$) can be added to increase nonlinearity.
\end{itemize}
The hyperbolic tangent ensures boundedness: $|\nu_i(s)| < \alpha$, so rotations remain within controlled limits. Orthogonality $(S_\nu^\top S_\nu = I)$ guarantees that the energy equation $s^\top(P-Q)s = f_p(s)^\top H_\gamma^{(p)} f_p(s)$ holds exactly.\newline

\subsubsection{Controlled open-loop stability}

A feature of the \texttt{NVDEx} family is the ability to create systems that are locally open-loop unstable, yet globally stabilizable by the optimal policy. This tests an algorithm's ability to handle unstable dynamics. This is achieved by tuning the linearization at the origin.

Let $A_0=H_\gamma^{-1/2}(P-Q)^{1/2}$ denote the linearization matrix at $s = 0$ (since $S_{\nu}(0) = I$). The drift simplifies to
\[
f_p(s) \approx H_\gamma^{-1/2} (P-Q)^{1/2} s =: A_0 s.
\]
The eigenvalues of $A_0$ determine local stability. Choosing $Q = \beta P$ with $\beta\in[0,1)$ gives:
\[
(P-Q)^{1/2} = (P-\beta P)^{1/2} = ((1-\beta)P)^{1/2} = \sqrt{1-\beta} P^{1/2}.
\]
We substitute this into the expression for $A_0$:
\[
A_0 = H_{\gamma}^{-1/2}(P-Q)^{1/2} = \sqrt{1-\beta}\big(H_{\gamma}^{-1/2}P^{1/2}\big).
\]
The spectral radius $\varrho(A_0)$ scales with this factor:
\[
\varrho(A_0) = \sqrt{1-\beta}\,\varrho\big(H_\gamma^{-1/2} P^{1/2}\big).
\]
Let $\varrho_{\text{base}} := \varrho(H_\gamma^{-1/2} P^{1/2})$. This is a fixed number determined by $H_\gamma$ and $P$. We tune $\beta$, reaching desired instability corresponding to a target spectral radius $\varrho_{\text{target}} > 1$. Then, 
\[
\varrho(A_0) = \sqrt{1-\beta}\varrho_{\text{base}} = \varrho_{\text{target}}.
\]
Solving for $\beta$ yields the following:
\[
\beta = 1 - \left(\frac{\varrho_{\text{target}}}{\varrho_{\text{base}}}\right)^2.
\]

If $\varrho_{\text{base}} \le \varrho_{\text{target}}$, then $\beta \le 0$, which is not allowed ($\beta$ must be in $[0,1)$). In this case, we first need to increase $\varrho_{\text{base}}$. Here, $\varrho_{\text{base}} = \varrho(H_\gamma^{-1/2} P^{1/2})$ depends on $H_\gamma$, which depends on $R$ (control cost). We select a cheaper control by scaling $R$ down with some factor smaller than $1$. This scales $H_\gamma$ up, increasing $\varrho_{\text{base}}$. We scale $R$ down until $\varrho_{\text{base}} > \varrho_{\text{target}}$, then compute $\beta$ as above.

Thus, choosing $\varrho_{\text{target}} > 1$ guarantees local open-loop instability, while the optimal policy $\pi_p^*$ stabilizes the system exactly to attain the optimal value $V^*$.\newline

\subsubsection{Optimal policy and value}

With the drift constructed as above, the optimal action is:
\[
a_p^* = -\gamma p\big(R + \gamma p^2 G^\top P G\big)^{-1} G^\top P f_p(s).
\]
As before, the additive Gaussian noise contributes only a constant offset to the value function, leaving the optimal action unchanged.\newline

\subsubsection{Difficulty knobs}

The family is parameterized by 
\[
\psi = (K, r_v, r_\omega, p, Q, R, P, \Sigma, \alpha, \kappa, \varrho_{\text{target}}).
\]
Key difficulty dimensions include:
\begin{itemize}
\item $K$: number of modules (scales state dimension);
\item $r_v, r_\omega$: integrator depths (increase dynamical complexity);
\item $p$: control authority (smaller $p$ makes stabilization harder);
\item $\alpha, \kappa$: nonlinearity gains in the coupling $\nu_i(s)$;
\item $\varrho_{\text{target}}$: desired open-loop instability margin.
\end{itemize}
By adjusting these parameters, we can generate benchmark instances ranging from nearly linear, easily controllable systems to highly nonlinear, unstable, high-dimensional challenges — all while preserving the certified optimal action $a_p^*$ and value function $V^*(s)$.

\subsection{Dataset construction}

From each family, we generate a dataset of validated instances by sampling parameters from specified distributions:
\begin{equation}
\mathcal{D}_N^{\text{(Arm)}} = \{ M_{\psi_i} : \psi_i \stackrel{\text{i.i.d.}}{\sim} P_{\Psi_{\text{Arm}}},\ i=1,\ldots,N_{\text{Arm}} \},
\end{equation}
and similarly for $\mathcal{D}_N^{\text{(NVDEx)}}$. The full benchmark dataset is the union:
\begin{equation}
\mathcal{D} = \mathcal{D}_{N_{\text{Arm}}}^{\text{(Arm)}} \cup \mathcal{D}_{N_{\text{NVDEx}}}^{\text{(NVDEx)}}.
\end{equation}

Each instance includes:
\begin{itemize}
\item Full parameter tuple $(Q,R,P,\gamma,\Sigma)$ and family metadata;
\item Deterministic seeds for initial-state schedules and noise streams;
\item Reference evaluators for $V^*$ and $a^*$;
\item Validation logs (SPD checks, Bellman residuals, boundedness tests).
\end{itemize}

The dataset supports paired evaluation under common random numbers (CRN) and enables ablations that separate algorithmic effects from protocol noise.

\section{Experimental Setup and Paired Evaluation}
We evaluate PPO, A2C, SAC, TD3, and DDPG (Stable-Baselines3) under identical action bounds and matched environment step budgets. On-policy methods are run on CPU; off-policy methods use GPU if available. Reward is $r(s,a)=-c(s,a)$. For each fixture, algorithms share identical initial-state schedules (ISS) and noise streams (CRN), and evaluation uses the analytic $(a^*,V_r^*)$, where $V_r^*$ is the optimal reward-to-go (analogous to cost-to-go $V^*$). For a start state $s_0$, the normalized optimality gap is
\begin{equation}\label{eq:optgap}
\mathrm{OptGap}(\pi;s_0)=\frac{V_r^\pi(s_0)-V_r^*(s_0)}{|V_r^*(s_0)|+\varepsilon},
\end{equation}
with $\varepsilon$ being a small constant to avoid division by 0. Discounted regret along the same CRN trajectory is
\begin{equation}\label{eq:regret}
\mathrm{Regret}_T(\pi;s_0)=\sum_{k=0}^{T-1}\gamma^k\Big(r(s_k,a^*_k)-r(s_k,\pi(s_k))\Big).
\end{equation}
We report means and 95\% confidence intervals via paired bootstrap over CRN trials.

\section{Results}
We report three regimes: \textit{serial\_random} (scheduled initial states), \textit{serial\_fixed} (repeated $s_0$), and \textit{NVDEx} (dynamic extension, longer horizon). These regimes vary occupancy measures and long-horizon conditioning while preserving the oracle $(a^*,V^*)$. As a result, protocol effects (coverage, distribution shift, estimator variance) and algorithmic effects (approximation error, exploration, bootstrapping bias) can be separated and interpreted in absolute units.\newline

\begin{table*}[t]
\caption{Overall performance per algorithm and experiment. Metrics are mean and 95\% CI via paired bootstrap under CRN.}
\label{tab:overall_performance}
\centering
\small
\begin{tabular}{llcccc}
\toprule
Experiment & Algorithm & OptGap (mean) & OptGap 95\% CI & Regret (mean) & Regret 95\% CI \\
\midrule
serial\_random & A2C  & -2.36e+03 & [-2.79e+03, -1.93e+03] & 8.24e+03 & [6.83e+03, 9.58e+03] \\
serial\_random & DDPG & -2.29e+03 & [-3.83e+03, -9.12e+02] & 9.28e+03 & [3.50e+03, 1.63e+04] \\
serial\_random & PPO  & -3.32e+02 & [-8.45e+02, -4.92e+01] & 1.51e+03 & [1.92e+02, 3.93e+03] \\
serial\_random & SAC  & -1.65e+02 & [-3.47e+02, -4.63e+01] & 6.90e+02 & [1.72e+02, 1.53e+03] \\
serial\_random & TD3  & -5.85e+01 & [-8.64e+01, -3.57e+01] & 2.14e+02 & [1.25e+02, 3.35e+02] \\
\midrule
serial\_fixed  & A2C  & -1.11e+03 & [-2.15e+03, -4.58e+02] & 5.78e+04 & [3.12e+04, 9.23e+04] \\
serial\_fixed  & DDPG & -6.48e+02 & [-1.34e+03, -2.51e+02] & 3.06e+04 & [1.61e+04, 5.22e+04] \\
serial\_fixed  & PPO  & -6.46e+01 & [-9.56e+01, -3.68e+01] & 4.08e+03 & [2.30e+03, 6.14e+03] \\
serial\_fixed  & SAC  & -1.51e+03 & [-2.44e+03, -8.00e+02] & 9.03e+04 & [5.18e+04, 1.33e+05] \\
serial\_fixed  & TD3  & -1.53e+03 & [-2.69e+03, -7.61e+02] & 8.31e+04 & [5.05e+04, 1.23e+05] \\
\midrule
NVDEx         & A2C  & -5.60e+03 & [-1.33e+04, -1.06e+03] & 5.94e+02 & [1.34e+02, 1.33e+03] \\
NVDEx         & DDPG & -1.46e+04 & [-3.13e+04, -3.37e+03] & 3.22e+03 & [4.76e+02, 7.39e+03] \\
NVDEx         & PPO  & -1.58e+01 & [-3.07e+01, -4.57e+00] & 2.97e+00 & [7.01e-01, 5.82e+00] \\
NVDEx         & SAC  & -7.76e+03 & [-1.56e+04, -1.15e+03] & 1.56e+03 & [1.13e+02, 3.13e+03] \\
NVDEx         & TD3  & -1.22e+04 & [-2.24e+04, -4.20e+03] & 2.78e+03 & [6.96e+02, 5.67e+03] \\
\bottomrule
\end{tabular}
\end{table*}

\begin{figure}[h]
    \centering
    \includegraphics[width=1.0\linewidth]{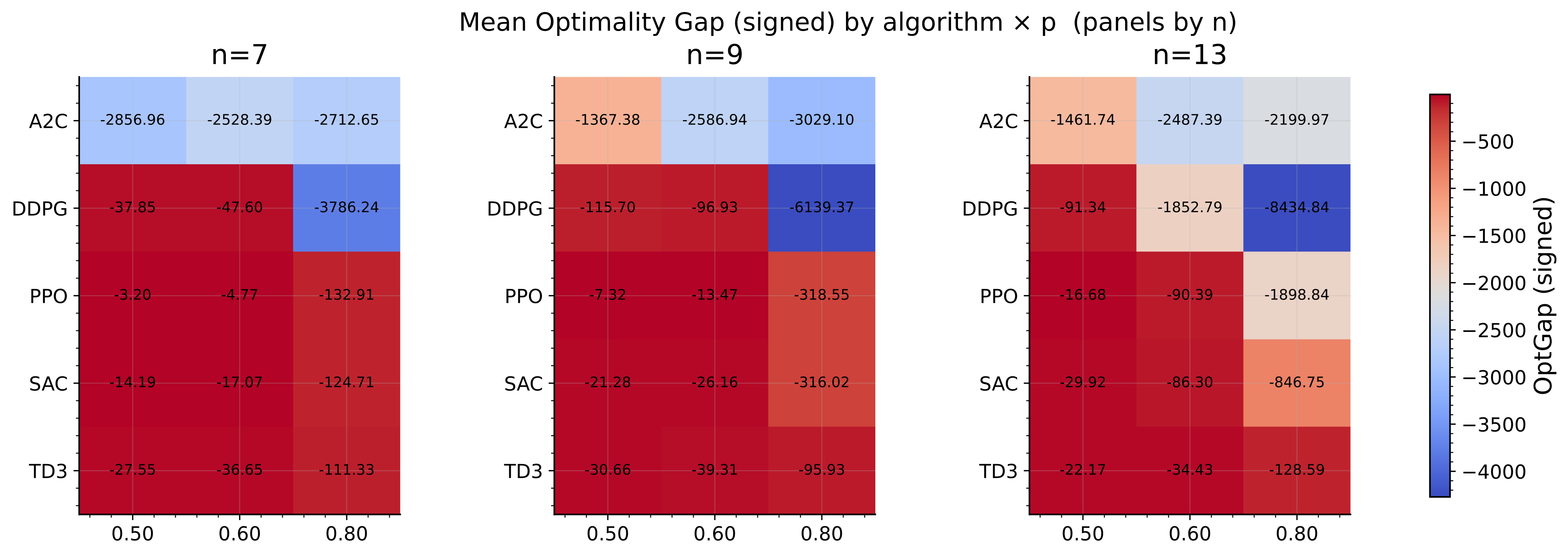}
    {\footnotesize (a) Optimality gap.}
\end{figure}
\begin{figure}[h]
    \centering
    \includegraphics[width=1\linewidth]{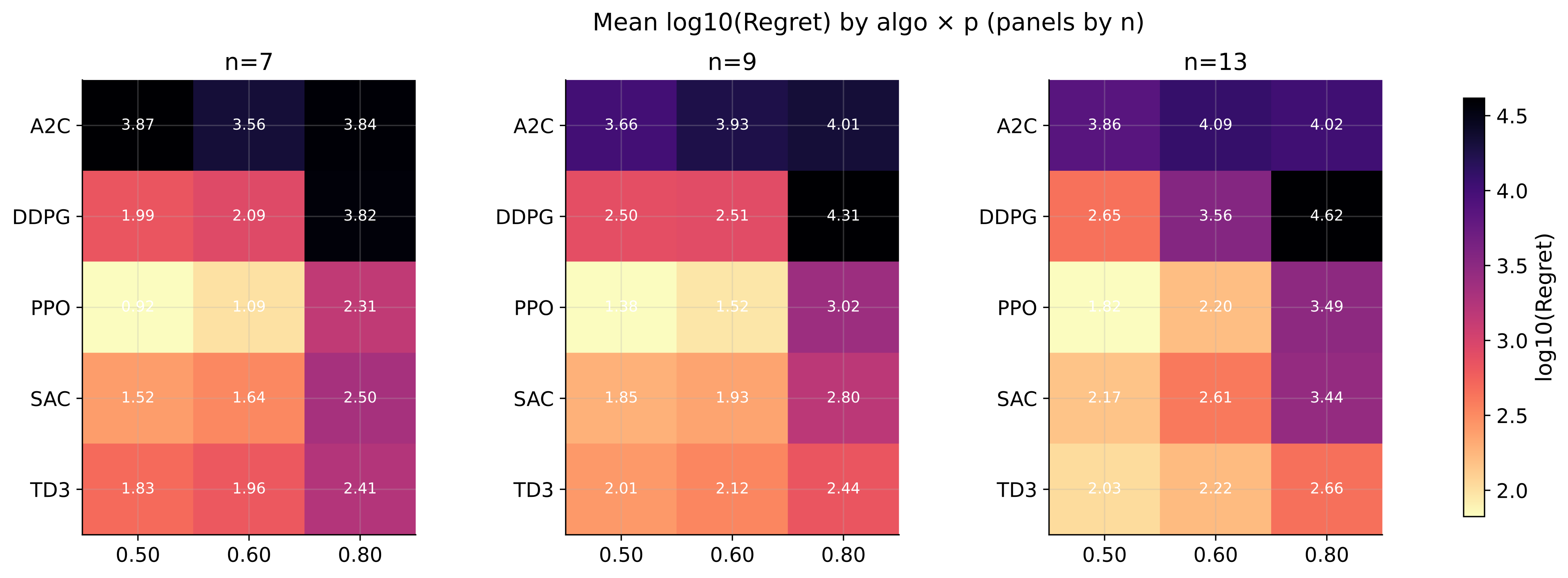}
    {\footnotesize (b) Regret ($\log_{10}$ scale).}
    \caption{Serial $n$-link planar arm (\texttt{ConverseArm}): absolute oracle-referenced performance (optimality gap and regret) across algorithms (rows) and control authority $p$ (columns), with separate panels for $n$.}
    \label{fig:heatmaps}
\end{figure}

\textit{Initialization protocol effects.} The performance divergence between \textit{serial\_random} and \textit{serial\_fixed} configurations underscores the critical role of state-distribution coverage in algorithmic efficacy. Randomized initial conditions facilitate diverse transition sampling in TD3 and SAC experience replay buffers, enhancing target estimate stability and data efficiency. However, it is important to note that in highly nonlinear and unstable dynamical systems, randomizing initial states may introduce substantial challenges, as algorithms must contend with widely varying system behaviors and potential instability regions that complicate policy learning. Conversely, fixed initial states produce narrow occupancy distributions where PPO’s constrained policy updates mitigate distributional shift, conferring advantages in stationary environments by focusing learning on a consistent operational regime.

\textit{Training budget considerations.} Disparate resource allocation across algorithms in the NVDEx domain warrants careful interpretation. PPO and A2C received substantially greater environmental interaction budgets, while off-policy methods incorporated initial warm-up periods (\texttt{learning\_starts}) that reduced their effective update windows. Although this allocation partially explains PPO’s dominance, the magnitude of performance differentials suggests additional contributing factors beyond mere step-count disparities.

\begin{figure}[h]
    \centering
    \includegraphics[width=1.0\linewidth]{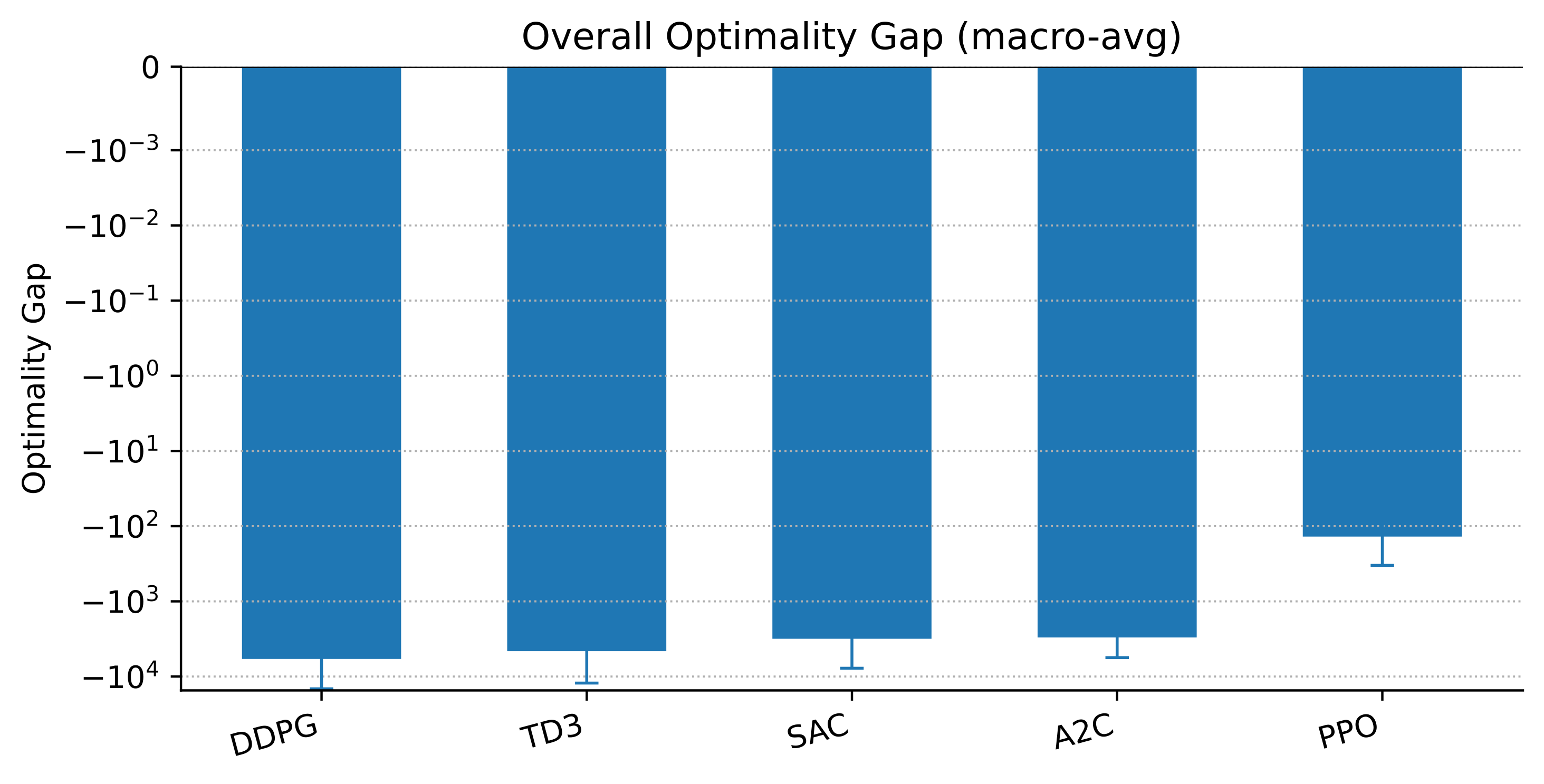}
    {\footnotesize (a) Macro-averaged optimality gaps of algorithms for all experiments (serial $n$-link planar arm and \textit{NVDEx}).}
\end{figure}
\begin{figure}[h]
    \centering
    \includegraphics[width=1\linewidth]{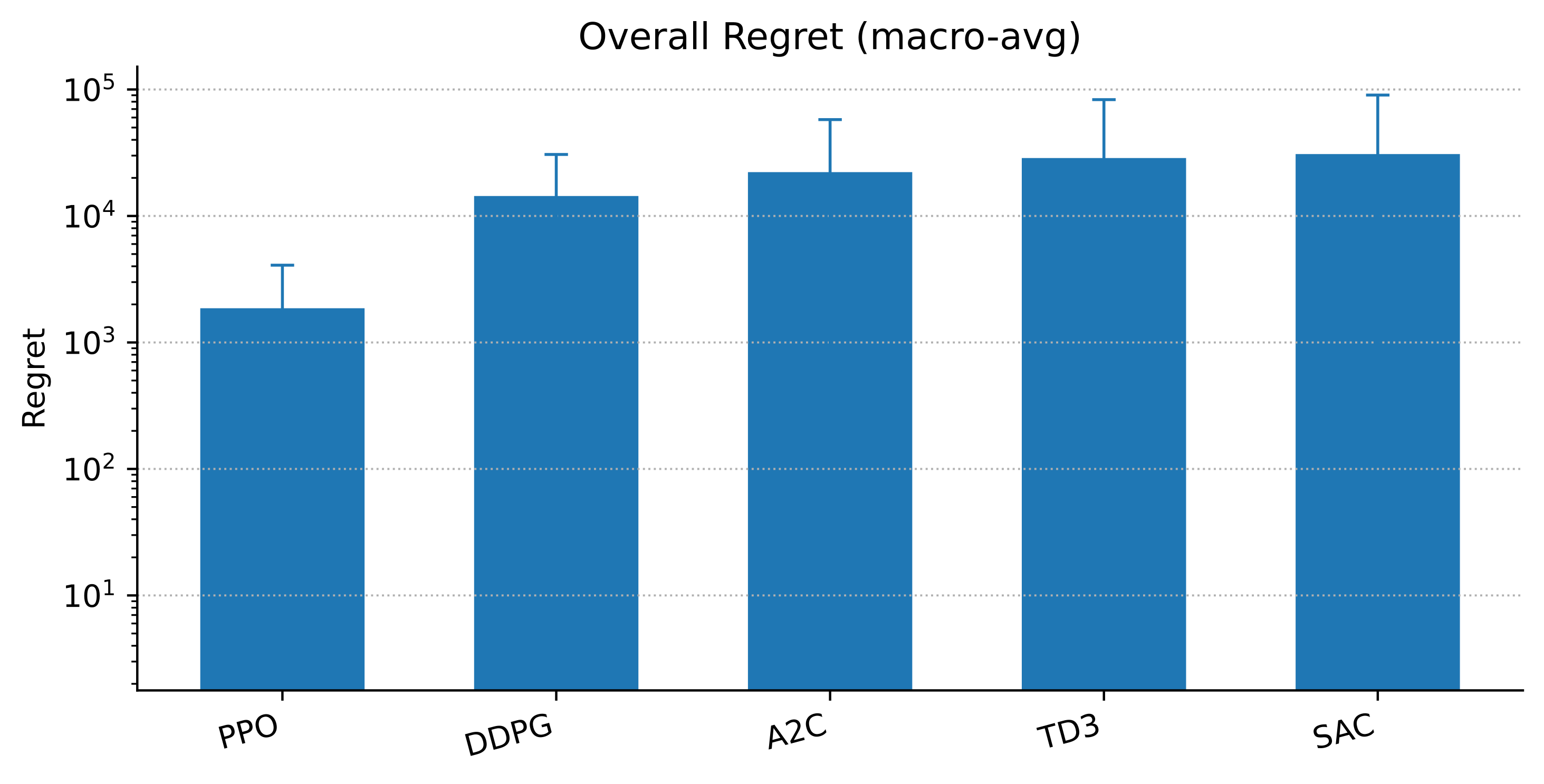}
    {\footnotesize (b) Macro-averaged regrets of algorithms for all experiments (serial $n$-link planar arm and \textit{NVDEx}).}
    \caption{Compact result summaries. Macro-averaged optimality gaps and
regrets across all experiments.}
    \label{fig:summary_bars}
\end{figure}

\begin{figure*}[h]
\centering
\begin{minipage}{1.0\textwidth}
  \centering
  \incfig[width=1.0\linewidth]{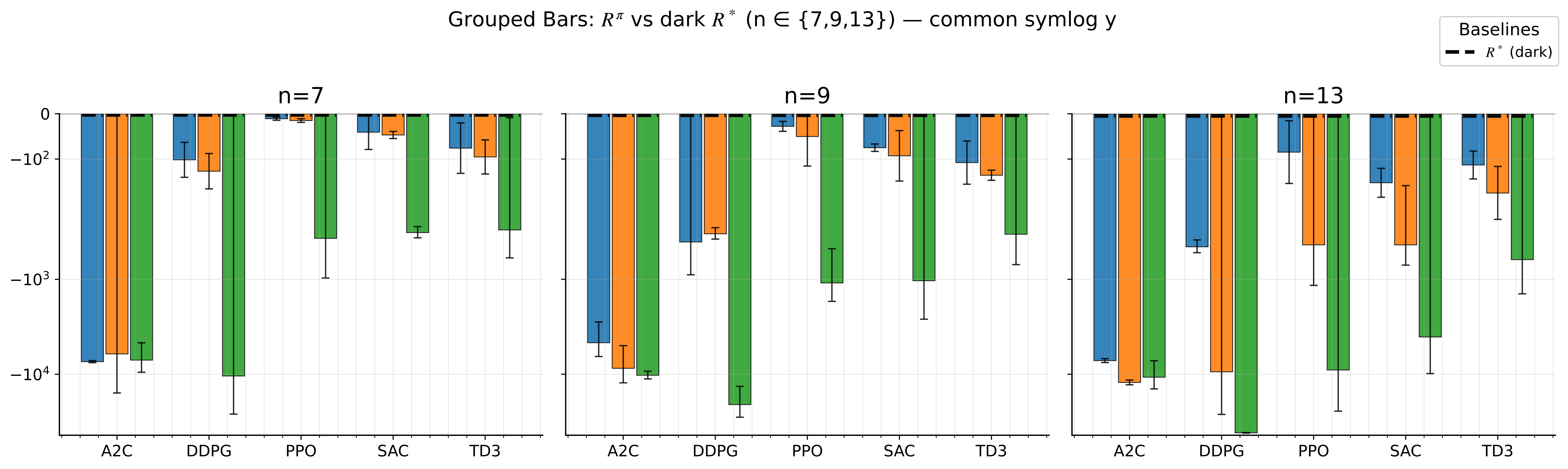}
  \vspace{-0.6em}
  \caption{Discounted reward bars ($r=-c$) with 95\% CIs under CRN. Colors correspond to $p \in \{0.5, 0.6, 0.8\}.$}
  \label{fig:groupedbars}
\end{minipage}
\end{figure*}

\textit{Dynamics and temporal horizon interactions.} The \textit{NVDEx} environment employs dynamic extension with extended temporal horizons ($T = 700$), generating stiff, feedback-sensitive closed-loop dynamics particularly at elevated ($p$, $\alpha$, $\kappa$) parameter values. These conditions amplify bootstrap estimation errors and critic approximation variance. PPO’s on-policy return estimates, regularized through clipping mechanisms, suggest superior robustness in such challenging regimes. This pattern recurs in the serial benchmark suite ($T = 512$ with state-dependent actuation), where off-policy methods excel under diverse state distributions (\textit{serial\_random}) while PPO prevails in narrow, stationary occupancies (\textit{serial\_fixed}).

\textit{Synthesis.} The empirical evidence suggests PPO is preferable for environments with stationary state distributions (\textit{serial\_fixed}) and complex dynamic systems with extended horizons (\textit{NVDEx}), while maintaining competitive performance under diverse initial conditions (\textit{serial\_random}). TD3 and SAC demonstrate particular strength in scenarios with rich replay buffer coverage. A2C and DDPG exhibit consistent underperformance across experimental conditions. These hierarchical performance relationships emerge from complex interactions among initialization protocols, effective training budgets, and environmental characteristics including temporal horizon and dynamic stiffness.\newline

Figures~\ref{fig:heatmaps},~\ref{fig:summary_bars}, and~\ref{fig:groupedbars} summarize performance. Heatmaps show how optimality gap and regret vary with control authority $p$ and dimension $n$ for \texttt{ConverseArm}; bar graphs aggregate between regimes, emphasizing broad trends. Darker shades correspond to higher regret (lower performance). As $n$ and $p$ increase, dynamics become more sensitive to control and approximation errors, making it harder for learned policies to match $\pi^*$. This visualization efficiently captures emergent patterns. Off-policy methods (TD3, SAC) dominate at high $p$ values and larger dimensions, and DDPG generally struggles due to ineffective exploration strategies, instability in value estimation, and sensitivity to reward scaling. On the other hand, PPO shows strong performance in moderate dimensions ($n \in \{7, 9\}$) and control authority ($p \leq 0.6$), benefiting from its robustness in policy, but struggles with high authority and dimensionality due to constrained trust regions and sample inefficiency. 

As $p$ increases (see Figs.~\ref{fig:heatmaps} and~\ref{fig:groupedbars}), regret consistently grows across algorithms may be explained by three related effects:

\begin{itemize}[leftmargin=*]
\item \textit{Accelerated oracle improvement}: The optimal policy $\pi^*$ becomes increasingly effective at dissipating energy in the $P$-norm as control authority grows. While the optimal reward $R^*$ decreases rapidly with increasing $p$, learned policies $R^\pi$ fail to improve at the same rate due to optimization and estimation limitations, resulting in widening regret $(R^\pi - R^*)$. This paradox arises because while stronger control makes the \emph{optimal} problem easier, it simultaneously amplifies sensitivity to policy imperfections, destabilizes learning dynamics, and increases function approximation complexity, making the \emph{learning} problem significantly harder.

\item \textit{Amplified sensitivity to imperfections}: The Jacobian of the closed-loop dynamics with respect to policy actions scales with $p$. Consequently, even minor policy errors, exploration noise, or value function estimation inaccuracies become dramatically amplified in the next-state distribution and state cost term $s^\top Q s$. The high control authority thus penalizes modest misestimation and introduces higher variance in bootstrapping targets.

\item \textit{Increased approximation complexity}: Both the discounted metric $H_\gamma^{(p)}(s)$ and the drift dynamics $f_p(s)$ exhibit a nonlinear dependence on $p$. The state-dependent actuation $g(s) = \cos\theta$ creates regions of reduced control effectiveness, while the system becomes increasingly stiff and reactive at higher $p$ values. These factors collectively destabilize gradients through long horizons ($T=512$), cause non-stationary replay distributions, and increase noise in GAE/TD targets, thereby slowing policy improvement.
\end{itemize}

\section{Discussion and Limitations}
The central advantage of our benchmarks is calibration: knowing $(a^*,V^*)$ disentangles algorithmic issues (exploration, approximation, bootstrapping) from protocol effects (initial states, noise, termination) -- impossible on standard suites. The continuation parameter $p$ varies control authority smoothly while preserving optimality, revealing when improved oracle performance coincides with increased learning sensitivity. Bounded nonlinear couplings via $S(\cdot)$ induce stiffness and state-dependent conditioning without sacrificing ground truth.

However, our assumptions impose limitations. The control-affine structure excludes general control dependencies. Additive i.i.d Gaussian noise misses heavy-tailed disturbances, heteroscedasticity, or temporal correlations. The QG specialization adds quadratic costs (excluding non-quadratic objectives, constraints) 
and linear optimal policies (excluding nonlinear controllers), with quadratic value functions restricting drifts via the energy equation.

These choices were deliberate, as they enable tractable construction of benchmarks with certified optima, but strong performance on our families does not guarantee robustness when these assumptions break. Extensions to broader noise, control, and cost models remain future work. 

\section{Conclusion}
We presented a framework that extends converse optimality to discounted stochastic control-affine systems with additive Gaussian noise and used the resulting construction to generate RL benchmarks with analytically known optimal policies and value functions. This method can also be used to benchmark classical control approaches. This enables reproducible, ground-truth evaluation via optimality gaps and regret under matched stochasticity and common random numbers. We release validated fixtures and code to support standardized comparisons. 

\section*{Artifacts and Reproducibility}
All fixtures, training/evaluation scripts, plotting code, and a dataset of other benchmarking systems are available at:
\href{https://github.com/converseoptimality/RL-Benchmarking}{github.com/converseoptimality/RL-Benchmarking}.

\bibliographystyle{IEEEtran}
\bibliography{bibliography}

\end{document}